\documentclass[]{article}
\usepackage[letterpaper]{geometry}
\usepackage{amta2022}
\usepackage{times}
\usepackage{url}
\usepackage{latexsym}
\usepackage{natbib}
\usepackage{layout}
\usepackage{graphicx}
\usepackage{subcaption}

\begin{document}

\title{\bf Efficient Machine Translation Corpus Generation}

\author{\name{\bf Kamer Ali Yuksel} \hfill  \addr{kamer@aixplain.com}\\
        \name{\bf Ahmet Gunduz} \hfill \addr{ahmet.gunduz@aixplain.com}\\
        \name{\bf Shreyas Sharma} \hfill \addr{shreyas.sharma@aixplain.com}\\
        \name{\bf Hassan Sawaf} \hfill \addr{hassan@aixplain.com}\\
        \addr{aiXplain Inc., 16535 Grant Bishop Lane, Los Gatos, CA 95032, USA}
}

\maketitle
\pagestyle{empty}

\begin{abstract}
This paper proposes an efficient and semi-automated method for human-in-the-loop post-editing for machine translation (MT) corpus generation. The method is based on online training of a custom MT quality estimation metric on-the-fly as linguists perform post-edits. The online estimator is used to prioritize worse hypotheses for post-editing, and auto-close best hypotheses without post-editing. This way, significant improvements can be achieved in the resulting quality of post-edits at a lower cost due to reduced human involvement. The trained estimator can also provide an online sanity check mechanism for post-edits and remove the need for additional linguists to review them or work on the same hypotheses. In this paper, the effect of prioritizing with the proposed method on the resulting MT corpus quality is presented versus scheduling hypotheses randomly. As demonstrated by experiments, the proposed method improves the lifecycle of MT models by focusing the linguist effort on production samples and hypotheses, which matter most for expanding MT corpora to be used for re-training them.
\end{abstract}

\section{Introduction}

Improving MT models requires continuously expanding their MT corpora for re-training cycles by post-editing their outputs on samples received from the production environment. Hence, the MT model lifecycle requires continuous human effort, which could scale and be more efficient by semi-automating it via machine-learning models trained by linguists. Those models can be used to select the maximally useful set of translations to store and post-edit by looking at what is challenging for an MT. They can upsample and prioritize translation outputs from where MTs are not performing well, and reduce costs by post-editing production translations intelligently. The continuous and interactive nature of the MT lifecycle provides the perfect ground for applying active-learning techniques in training those machine-learning models for semi-automation. Custom translation quality or post-editing effort estimation models trained on-the-fly as linguists post-edit translations can be used to prioritize samples accumulating from the model inferences in the production environment. The trained estimators enable to focus the linguist effort on the most challenging samples for the MT model requiring the most post-edits, which are also the most valuable to check for evaluating the MT model quality by humans.

In the WMT20 Metrics Shared Task \citep{mathur2020results}, participants were asked to score MT outputs in the WMT20 News Translation Task with automatic metrics, and four referenceless metrics were submitted. Those metrics (OpenKiwi-BERT, OpenKiwi-XLMR, YISI-2, COMET-QE) use bilingual mappings of the contextual embeddings extracted from pre-trained or fine-tuned language models (like XLM-RoBERTa) to evaluate the cross-lingual lexical semantic similarity between the input and MT output. However, it has been seen that those metrics generally struggle to score human translations against machine translations reliably except for COMET-QE \citep{rei2020comet}, which was the only reference-free metric that was able to differentiate human translations from MT. \citet{freitag2021experts} carried out an MQM research study by scoring the outputs of top systems from the WMT20 Metrics Shared Task in two language pairs using annotations provided by professional translators with access to the entire document context. Their study shows that crowd-worker human evaluations (as conducted by WMT) have a low correlation with MQM, and the resulting system-level rankings are quite different; and questioned previous conclusions based on crowd-worker human evaluation, especially for high-quality MT. Most importantly, they also found that automatic metrics based on pre-trained embeddings can outperform human crowd workers. This was a clear indication that machine learning models trained over crowd-sourced human-evaluations can reach a higher generalization performance than individual evaluators; thus, they can be used for sanity-checking.

In the WMT21 Metrics Shared Task \citep{freitag2021results}, contrary to the previous year, they have also acquired their human ratings based on expert-based human evaluation, which has shown to be more reliable, via MQM; and were able to evaluate all metrics on two different domains (news and TED talks) using translations of the same MT systems. It has been found that reference-free metrics (in particular COMET-QE and OpenKiwi) perform very well in scoring human translations but not as well with MT outputs. They are also relatively good at rating human translations at the segment-level while being competitive against their reference-based counterparts in system-level evaluation. REGEMT \citep{vstefanik2021regressive} was a new reference-free metric of WMT21, which was created as an ensemble of other selected metrics of surface, syntactic and semantic-level similarity as input features to a regression model that estimates a quality assessment. It used the following input features: Source length, Target length, Contextual SCM, Contextual WMD, BERTScore, Prism, and Compositionality. The ensembling can allow customization and continual learning of their quality estimation metrics. cushLEPOR \citep{han2021cushlepor} customized hLEPOR metric by hyper-tuning its weighting-parameters to better agree with professional human evaluations, including on MQM and pSQM scores; and achieved competitive results against quality estimation metrics based on pre-trained neural models measuring cross-lingual lexical semantic similarity, at a much higher cost.

The primary contributions of this paper are three-fold: (1) proposing a new architecture for managing MT model production lifecycle in a cost-efficient and scalable semi-automated way (2) demonstrating the effective use of referenceless metrics for training dataset building (via post-editing), and human evaluation processes of this lifecycle (3) active-learning of custom referenceless metrics as machine learning targets are collected from the human-annotators. To the best of our knowledge, there is no previous work that studies how to employ quality estimation metrics to improve the production lifecycle of MT systems by prioritizing the incoming translations to be evaluated or post-edited. Furthermore, none of the previous work in the literature also studied how custom quality metrics can be trained on-the-fly in an active-learning fashion. Finally, none of them also reported how that would affect the quality of human  post-edits by semi-automation at different human-involvement levels. The comparison between random prioritization and the personalized metric is provided in the experiments section, not only by scoring accuracy but also by their effect on reference building by post-editing.

\section{Related Work}

There have been previous attempts to use active-learning for more efficient corpus extension for MT, but those were using model-free (based on diversity) and (neural MT) model-based uncertainty sampling methods. \cite{peris2018active} used active-learning for interactive MT where they have selected hypotheses that are worth being supervised by human agents by exploiting the attention mechanism of a neural MT as a measure of uncertainty. \cite{zeng2019empirical} used paraphrastic embeddings from unsupervised pre-training to sample diverse sentences for active-learning in MT. They have also proposed an alternative using information-loss during bi-directional translation. \cite{hu2021phrase} also performed uncertainty-based active-learning for fine-tuning MTs by selecting phrases for translation rather than translating entire sentences. In this work, instead of using uncertainty-based proxy-measures for the difficulty of samples for the MT, a reference-free quality-estimator is trained online with MT errors measured by each post-edit. This allows the proposed method not just to select the most challenging sample for post-editing but also to select which hypothesis should be used for post-editing when multiple of them are present. Furthermore, the trained quality estimation model provides a mechanism for sanity-checking or gamifying the post-editing activity of linguists; and also enables augmenting the training corpus by using high-quality hypotheses as pseudo-references for self-training.

\section{Methodology}

Post-editing should be performed continuously on the translations of MT models in-production for: (1) deciding when an MT model is good enough for deployment, (2) deciding if a new MT is better than its version in-deployment, (3) deciding when a deployed MT model needs to be re-trained, (4) obtaining references to use for re-training MT models with extended corpora. This work aimed to improve and scale all those manual processes for managing the lifecycle of MT models in-production for their continual improvement, evaluation, and debugging. Estimating the difficulty of translations for an MT can be helpful in deciding which of them to post-edit. This way, the manual labor in post-editing can be reduced; while increasing its effectiveness by upsampling challenging translations where the MT is estimated not to perform well.

In this work, MT corpus generation is conducted more effectively by training a machine-learning (ML) model iteratively with each post-edit from the linguists. This ML model is used to efficiently generate MT corpora by prioritizing post-edit efforts of human translators and providing them real-time feedback through model predictions of the post-edits. The ML model can simultaneously be used for performing sanity checks on linguists by checking the discrepancies between each and its own decisions. MT hypotheses that are efficiently and semi-automatically post-edited, can be used as training corpora to re-train or fine-tune MTs, or as validation corpora to benchmark them. Many MT vendors allow their customers to customize their MTs to their application for better performance when a custom corpus is available with references. As a result of selecting and post-editing translations from production systems, one also obtains references necessary to automatically score them with industry-standard MT scoring metrics.

\begin{figure}[bhp]
\includegraphics[width=\columnwidth]{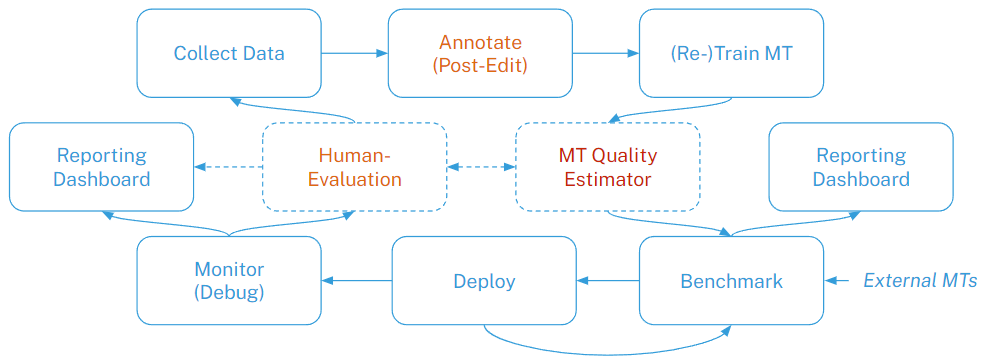}
\caption{The architecture of the proposed method for efficient MT corpus generation.}
\label{fig:arch}
\end{figure}

\newpage 

The architecture of the proposed method is illustrated in Figure \ref{fig:arch} where a reference-free custom MT quality estimator, which is trained online by obtained post-edits iteratively, is the central component that drives the continuous post-editing process in MT model production lifecycle efficiently by prioritizing it. The obtained post-edits can be treated as ground-truth references for benchmarking with reference-based scores on translations coming from the deployment, and later for re-training a new version of the MT model that will be shadow-tested against the previous version in-production environment. External MTs can optionally be used in the process not just for comparing against but also for providing alternative translations selected for post-editing when less linguist effort is estimated. They can also enable round-trip translations of source texts for the data-augmentation of training samples for the estimator. 

The estimator trained with online AutoML functionalities in FLAML framework \citep{wang2021flaml}, uses fine-tuned cross-lingual embeddings of COMET-QE score as features, alongside many other linguistic ones extracted with Stanza \citep{qi2020stanza} from source texts and their translations: the number of tokens, characters, and the average word length of sentences; the frequency of Part-of-Speech and Named Entity Recognition labels, and the frequency of morphological features. The differences in values of linguistic features and COMET-QE embeddings between source texts and translations, cosine distance of their COMET-QE embeddings, and the pointwise product of COMET-QE embeddings of source texts and their translations are also included as features. When the source or target language is English, an additional 250 linguistic (syntax, semantics, discourse, and readability) features are extracted with LingFeat library \citep{lee-etal-2021-pushing} in SpaCy. COMET-DA (the reference-based version of COMET metric trained on Direct Assessments) and Translation-Error-Rate (TER) metrics in-between MT hypotheses and their respective post-edits are attempted as regression targets, with mean-squared error as the training objective. After the model update on each post-edit, the next sample with the lowest estimated COMET-DA or highest estimated TER is prioritized for the linguist post-editing.

\begin{figure*}[t]
  \centering
  \begin{minipage}{\columnwidth}
    \begin{subfigure}[b]{0.5\columnwidth}
      \includegraphics[width=\linewidth]{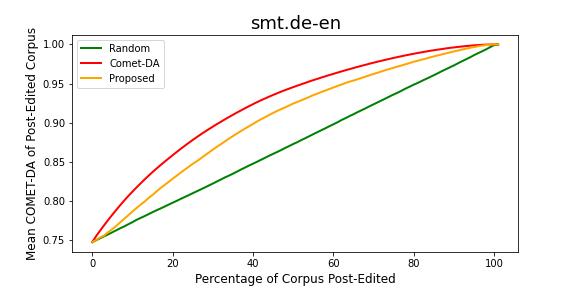}
    \end{subfigure}%
     \begin{subfigure}[b]{0.5\columnwidth}
      \includegraphics[width=\linewidth]{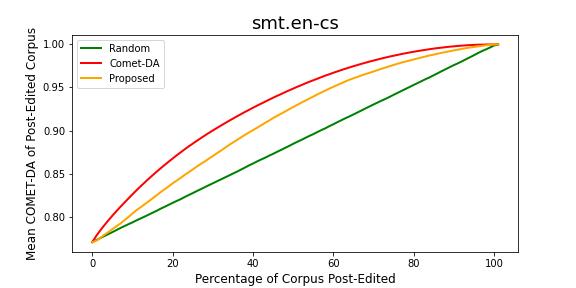}
    \end{subfigure}
     \begin{subfigure}[b]{0.5\columnwidth}
      \includegraphics[width=\linewidth]{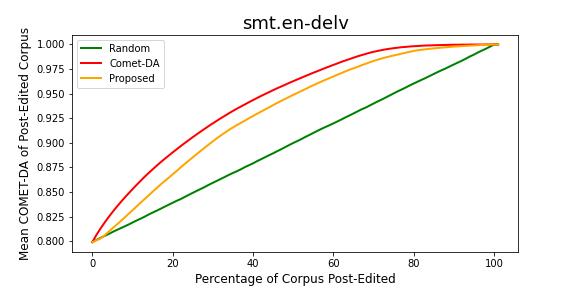}
    \end{subfigure}%
     \begin{subfigure}[b]{0.5\columnwidth}
      \includegraphics[width=\linewidth]{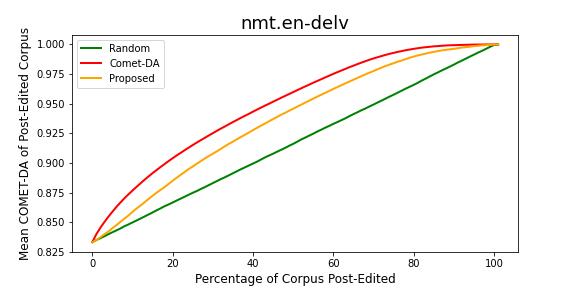}
    \end{subfigure}
  \end{minipage}%
  \caption{The corpus quality (measured by COMET-DA) achieved when hypotheses are ranked by COMET-DA estimator (yellow), randomly (green), and COMET-DA ground-truth (red).}
  \label{fig:comet_da}
\end{figure*}

\begin{table*}[!b]
\centering
\begin{tabular}{l|r|r|r|r|r|r}
\hline
Sub-corpus & Samples & MAE & MSE & Spearman $\rho$ & Pearson \emph{r} & Kendall $\tau$ \\
\hline
\textbf{SMT de-en}    & 43000 & 0.13 & 0.03 & 0.53 & 0.46 & 0.38 \\
\textbf{SMT en-cs}    & 43000 & 0.12 & 0.02 & 0.52 & 0.46 & 0.36 \\
\textbf{SMT en-de,lv} & 46738 & 0.10 & 0.02 & 0.65 & 0.55 & 0.46 \\
\textbf{NMT en-de,lv} & 46738 & 0.08 & 0.01 & 0.48 & 0.44 & 0.34 \\
\hline
\end{tabular}
\caption{The regression and ranking performance of the online estimator on four sub-corpora.}
\label{tab:reg}
\end{table*}

\begin{table*}[]
\resizebox{\textwidth}{!}{%
\begin{tabular}{l|rr|rr|rr|rr|rr|rr|rr}
\hline
\textbf{SMT de-en} & 20\% & $\Delta$ & 30\% & $\Delta$ & 40\% & $\Delta$ & 50\% & $\Delta$ & 60\% & $\Delta$ & 70\% & $\Delta$ & 80\% & $\Delta$ \\ \hline
Random & 0.80 & 0.00\% & 0.82 & 0.00\% & 0.85 & 0.00\% & 0.87 & 0.00\% & 0.90 & 0.00\% & 0.92 & 0.00\% & 0.95 & 0.00\% \\
Proposed & 0.83 & +3.85\% & 0.87 & +5.27\% & 0.90 & \textbf{+6.00\%} & 0.92 & +5.92\% & 0.94 & +5.23\% & 0.96 & +4.20\% & 0.98 & +3.09\% \\ \hline
\textbf{SMT en-cs} & 20\% & $\Delta$ & 30\% & $\Delta$ & 40\% & $\Delta$ & 50\% & $\Delta$ & 60\% & $\Delta$ & 70\% & $\Delta$ & 80\% & $\Delta$ \\ \hline
Random & 0.82 & 0.00\% & 0.84 & 0.00\% & 0.86 & 0.00\% & 0.88 & 0.00\% & 0.91 & 0.00\% & 0.93 & 0.00\% & 0.95 & 0.00\% \\
Proposed & 0.84 & +2.78\% & 0.87 & +3.78\% & 0.90 & +4.50\% & 0.93 & \textbf{+4.83\%} & 0.95 & +4.79\% & 0.97 & +4.12\% & 0.98 & +3.11\% \\ \hline
\textbf{SMT en-de,lv} & 20\% & $\Delta$ & 30\% & 30\% & 40\% & $\Delta$ & 50\% & $\Delta$ & 60\% & $\Delta$ & 70\% & $\Delta$ & 80\% & $\Delta$ \\ \hline
Random & 0.84 & 0.00\% & 0.86 & 0.00\% & 0.88 & 0.00\% & 0.90 & 0.00\% & 0.92 & 0.00\% & 0.94 & 0.00\% & 0.96 & 0.00\% \\
Proposed & 0.87 & +3.44\% & 0.90 & +4.90\% & 0.93 & \textbf{+5.47\%} & 0.95 & +5.43\% & 0.97 & +5.19\% & 0.98 & +4.61\% & 0.99 & +3.45\% \\ \hline
\textbf{NMT en-de,lv} & 20\% & $\Delta$ & 30\% & $\Delta$ & 40\% & $\Delta$ & 50\% & $\Delta$ & 60\% & $\Delta$ & 70\% & $\Delta$ & 80\% & $\Delta$ \\ \hline
Random & 0.87 & 0.00\% & 0.88 & 0.00\% & 0.90 & 0.00\% & 0.92 & 0.00\% & 0.93 & 0.00\% & 0.95 & 0.00\% & 0.97 & 0.00\% \\
Proposed & 0.88 & +2.12\% & 0.91 & +2.81\% & 0.93 & +3.14\% & 0.95 & \textbf{+3.24\%} & 0.96 & +3.16\% & 0.98 & +2.98\% & 0.99 & +2.47\% \\ \hline
\end{tabular}%
}
\caption{The mean corpus quality (in COMET-DA) when prioritized by the proposed COMET-DA estimator versus random-prioritization baseline on different percentages of post-editing.}
\label{tab:comet-da}
\end{table*}

\begin{table*}[]
\resizebox{\textwidth}{!}{%
\begin{tabular}{l|rr|rr|rr|rr|rr|rr|rr}
\hline
\textbf{SMT de-en} & 20\% & $\Delta$ & 30\% & $\Delta$ & 40\% & $\Delta$ & 50\% & $\Delta$ & 60\% & $\Delta$ & 70\% & $\Delta$ & 80\% & $\Delta$ \\ \hline
Random      & 60.13 & 0.00\%                    & 64.76 & 0.00\% & 69.72 & 0.00\% & 74.72 & 0.00\% & 79.42 & 0.00\% & 85.71 & 0.00\% & 89.84 & 0.00\% \\
Proposed     & 62.98 & 4.74\%                    & 69.05 & 6.62\% & 74.92 & 7.45\% & 80.19 & 7.32\% & 85.72 & \textbf{7.94\%} & 90.43 & 6.76\% & 94.21 & 4.86\% \\ \hline
\textbf{SMT en-cs}   & 20\%  & $\Delta$ & 30\%  & $\Delta$      & 40\%  & $\Delta$      & 50\%  & $\Delta$      & 60\%  & $\Delta$      & 70\%  & $\Delta$      & 80\%  & $\Delta$      \\ \hline
Random                & 57.03 & 0.00\%                    & 62.36 & 0.00\% & 67.70 & 0.00\% & 73.05 & 0.00\% & 78.38 & 0.00\% & 83.72 & 0.00\% & 89.11 & 0.00\% \\
Proposed              & 59.26 & 3.90\%                    & 65.81 & 5.53\% & 71.70 & 5.90\% & 77.46 & \textbf{6.03\%} & 82.87 & 5.74\% & 87.90 & 4.99\% & 92.46 & 3.77\% \\ \hline
\textbf{SMT en-de,lv} & 20\%  & $\Delta$ & 30\%  & 30\%   & 40\%  & $\Delta$      & 50\%  & $\Delta$      & 60\%  & $\Delta$      & 70\%  & $\Delta$      & 80\%  & $\Delta$      \\ \hline
Random       & 60.18 & 0.00\%                    & 65.08 & 0.00\% & 70.02 & 0.00\% & 74.89 & 0.00\% & 79.74 & 0.00\% & 84.69 & 0.00\% & 89.73 & 0.00\% \\
Proposed     & 63.50 & 5.53\%                    & 69.66 & 7.03\% & 75.41 & 7.71\% & 80.92 & \textbf{8.05\%} & 85.61 & 7.35\% & 90.10 & 6.39\% & 94.24 & 5.03\% \\ \hline
\textbf{NMT en-de,lv} & 20\%  & $\Delta$ & 30\%  & $\Delta$      & 40\%  & $\Delta$      & 50\%  & $\Delta$      & 60\%  & $\Delta$      & 70\%  & $\Delta$      & 80\%  & $\Delta$      \\ \hline
Random                & 61.78 & 0.00\%                    & 60.50 & 0.00\% & 71.24 & 0.00\% & 75.93 & 0.00\% & 80.66 & 0.00\% & 86.37 & 0.00\% & 90.25 & 0.00\% \\
Proposed              & 64.08 & 3.73\%                    & 70.09 & 5.40\% & 75.52 & 6.00\% & 80.56 & \textbf{6.09\%} & 85.20 & 5.62\% & 89.51 & 4.85\% & 93.54 & 3.65\% \\ \hline
\end{tabular}%
}
\caption{The mean corpus quality (measured by 100-TER) when prioritized by the proposed TER estimator versus random-prioritization baseline on various percentages of post-editing.}
\label{tab:reverse-ter}
\end{table*}

\section{Experiments}

The QT21 corpus \citep{11372/LRT-2390}, a publicly available dataset containing industry-generated sentences from information technology and life sciences domains, has been used in the experiments. The corpus contained 179K tuples of source and their respective reference sentences for which an MT hypothesis, either from a statistical or a neural MT model, and its post-edit is also provided. Some (43K) of the tuples in the dataset were for German-English (de-en) language-pair; whereas the remaining 136K was translations from English into one of the target languages (Czech, German, Latvian) available. The experiments were conducted for the hypotheses of statistical and neural MTs (SMT and NMT, respectively) individually on four sub-corpora. Table \ref{tab:reg} shows the size of each sub-corpus and the performance of the COMET-DA estimator regarding blind predictions collected during online training (before training with each sample). It can be seen that the estimator achieves a high ranking-correlation with the regression target, which demonstrates its capability in prioritizing post-edits regarding how critical they are.

The experimental results are presented for MT quality estimators trained with COMET-DA and TER  targets. The estimators trained online with each post-edit are used to re-prioritize the post-editing queue after each learning step. The mean quality or translation error of the corpus is calculated with ground-truth references available and logged after each post-edit. The MT models  are not re-trained with generated corpora during the course of post-editing. It has been observed that the proposed method reaches a better corpus quality with fewer post-edits in all of the datasets as shown in the following results. Table \ref{tab:comet-da} and \ref{tab:reverse-ter} shows the percentage gain of the proposed method on each dataset against ranking randomly in terms of COMET-DA and TER respectively. 
It can be observed that the mean corpus quality obtains the highest gain (6-8\% in inverse-TER) versus ranking randomly when around half of the corpus is post-edited, where the mean TER of the corpus would be lower up to 20\% than ranking-randomly for post-editing.

It can be assumed that due to limited resources, only a portion of the whole dataset would be post-edited by linguists in many real-world cases. Based on the demonstrated experimental results, one can conclude that active-learning to schedule post-edits leads to more efficient use of linguists than ranking hypotheses randomly for post-editing. As shown also in Figure \ref{fig:comet_da}, the Pareto-optimal corpus quality is achieved when 40-50\% of those hypotheses are post-edited by de-prioritizing ones where post-editing would not lead to a significant improvement. The mean corpus quality indicated by COMET-DA already reaches up to 95\% when half of the hypotheses are post-edited with up to half error (inverse COMET-DA) than  ranking randomly. As the reduction of error is more significant in COMET-DA than in TER, it can be said that humans are less sensitive in their Direct Assessments to MT errors than automatic metrics when the hypothesis quality is already reasonably high. It has also been observed the proposed method contributes more to the corpus quality when prioritizing SMT hypotheses where it achieves higher ranking-correlation despite reaching better regression performance for NMT. This is probably because the qualities of SMT hypotheses are more heterogeneous than NMT.

\section{Discussion}

Since the proposed method depends on the embeddings from XLM-RoBERTa model, it is limited to 100 languages that the model was pre-trained with. However, the experimental results indicate that the proposed method is able to generalize to the languages (like Czech and Latvian), which have not been employed in the fine-tuning of that encoder for COMET-QE model. The online training with those language-agnostic embeddings helps the estimator in quickly adapting to the unseen languages or training multi-language estimators as it has been done in this work for German and Latvian by providing the target language as an input feature. 

Despite the effect of the proposed method on the corpus quality studied in this work, the resulting effect on the MT model performance after re-training is not measured quantitatively and that is planned to be part of future-work; but it can be assumed to also improve with the better corpus quality where the difficult samples for the MT are prioritized to extend its training corpus. Moreover, the effect of augmenting the MT corpus, by using high-quality MT hypotheses for production samples as pseudo-references, on the resulting MT performance should also be studied. In addition, the contribution of MT re-training iterations on further improving the overall MT corpus quality by updating these pseudo-references can also be measured in future-work. Finally, the combination of the proposed method with diversity-based active-learning techniques (especially using extracted embeddings) will also be studied in future-work, and the experiments will also be extended to compare with those uncertainty-based techniques.


\section{Conclusion}
Efficiently obtaining references for MT corpora continuously accumulated from production is crucial for improving MT models. MT corpus generation is a costly manual process, and its efficiency and scalability can be significantly improved by training an online ML model that prioritizes the post-editing workload of linguists with high accuracy. In this work, it has been shown that the post-editing process can be improved by prioritizing the samples that need it the most - the ones from which MTs would learn most when re-trained with obtained references. It can be expected that prioritizing the most challenging production samples for corpus generation would also lead to better hypotheses on the remaining samples when MTs are re-trained. The trained estimator can also be helpful in sanity-checking the post-editing performance of each linguist online without the need of a reviewer or a duplicated effort of post-editing to create references. It can also be used to give them feedback to gamify their post-editing process. When hypotheses from multiple MTs are present for each source-text, the estimator can also be used to pre-select the best hypothesis to post-edit, and further increase the linguist efficiency. Finally, it can also be used to prioritize the post-edits of linguists for a manual reviewing process.

\small

\bibliographystyle{apalike}
\bibliography{amta2022}

\begin{thebibliography}{}

\bibitem[Freitag et~al., 2021a]{freitag2021experts}
Freitag, M., Foster, G., Grangier, D., Ratnakar, V., Tan, Q., and Macherey, W.
  (2021a).
\newblock Experts, errors, and context: A large-scale study of human evaluation
  for machine translation.
\newblock {\em Transactions of the Association for Computational Linguistics},
  9:1460--1474.

\bibitem[Freitag et~al., 2021b]{freitag2021results}
Freitag, M., Rei, R., Mathur, N., Lo, C.-k., Stewart, C., Foster, G., Lavie,
  A., and Bojar, O. (2021b).
\newblock Results of the wmt21 metrics shared task: Evaluating metrics with
  expert-based human evaluations on ted and news domain.
\newblock In {\em Proceedings of the Sixth Conference on Machine Translation},
  pages 733--774.

\bibitem[Han et~al., 2021]{han2021cushlepor}
Han, L., Sorokina, I., Erofeev, G., and Gladkoff, S. (2021).
\newblock cushlepor: customising hlepor metric using optuna for higher
  agreement with human judgments or pre-trained language model labse.
\newblock In {\em Proceedings of the Sixth Conference on Machine Translation},
  pages 1014--1023.

\bibitem[Hu and Neubig, 2021]{hu2021phrase}
Hu, J. and Neubig, G. (2021).
\newblock Phrase-level active learning for neural machine translation.
\newblock {\em arXiv preprint arXiv:2106.11375}.

\bibitem[Lee et~al., 2021]{lee-etal-2021-pushing}
Lee, B.~W., Jang, Y.~S., and Lee, J. (2021).
\newblock Pushing on text readability assessment: A transformer meets
  handcrafted linguistic features.
\newblock In {\em Proceedings of the 2021 Conference on Empirical Methods in
  Natural Language Processing}, pages 10669--10686, Online and Punta Cana,
  Dominican Republic. Association for Computational Linguistics.

\bibitem[Mathur et~al., 2020]{mathur2020results}
Mathur, N., Wei, J., Freitag, M., Ma, Q., and Bojar, O. (2020).
\newblock Results of the wmt20 metrics shared task.
\newblock In {\em Proceedings of the Fifth Conference on Machine Translation},
  pages 688--725.

\bibitem[Peris and Casacuberta, 2018]{peris2018active}
Peris, {\'A}. and Casacuberta, F. (2018).
\newblock Active learning for interactive neural machine translation of data
  streams.
\newblock {\em arXiv preprint arXiv:1807.11243}.

\bibitem[Qi et~al., 2020]{qi2020stanza}
Qi, P., Zhang, Y., Zhang, Y., Bolton, J., and Manning, C.~D. (2020).
\newblock Stanza: A {Python} natural language processing toolkit for many human
  languages.
\newblock In {\em Proceedings of the 58th Annual Meeting of the Association for
  Computational Linguistics: System Demonstrations}.

\bibitem[Rei et~al., 2020]{rei2020comet}
Rei, R., Stewart, C., Farinha, A.~C., and Lavie, A. (2020).
\newblock Comet: A neural framework for mt evaluation.
\newblock {\em arXiv preprint arXiv:2009.09025}.

\bibitem[Specia, 2017]{11372/LRT-2390}
Specia, L. (2017).
\newblock {QT21} data.
\newblock {LINDAT}/{CLARIAH}-{CZ} digital library at the Institute of Formal
  and Applied Linguistics ({{\'U}FAL}), Faculty of Mathematics and Physics,
  Charles University.

\bibitem[{\v{S}}tef{\'a}nik et~al., 2021]{vstefanik2021regressive}
{\v{S}}tef{\'a}nik, M., Novotn{\`y}, V., and Sojka, P. (2021).
\newblock Regressive ensemble for machine translation quality evaluation.
\newblock {\em arXiv preprint arXiv:2109.07242}.

\bibitem[Wang et~al., 2021]{wang2021flaml}
Wang, C., Wu, Q., Weimer, M., and Zhu, E. (2021).
\newblock Flaml: a fast and lightweight automl library.
\newblock {\em Proceedings of Machine Learning and Systems}, 3:434--447.

\bibitem[Zeng et~al., 2019]{zeng2019empirical}
Zeng, X., Garg, S., Chatterjee, R., Nallasamy, U., and Paulik, M. (2019).
\newblock Empirical evaluation of active learning techniques for neural mt.
\newblock In {\em Proceedings of the 2nd Workshop on Deep Learning Approaches
  for Low-Resource NLP (DeepLo 2019)}, pages 84--93.

\end{thebibliography}

\end{document}